\documentclass{bmvc2k}

\title{A self-supervised cyclic neural-analytic approach for novel view synthesis and 3D reconstruction}

\addauthor{Drago\c{s} Costea}{dragos.costea@upb.ro}{1}
\addauthor{Alina Marcu}{alina.marcu@upb.ro}{2}
\addauthor{Marius Leordeanu}{marius.leordeanu@upb.ro}{1,2,3}

\addinstitution{
National University of Science and Technology "Politehnica" Bucharest\\
}
\addinstitution{
"Simion Stoilow" Institute of Mathematics of the Romanian \\ Academy
}
\addinstitution{
NORCE Norwegian Research \\ Centre AS
}

\runninghead{Costea, Marcu, Leordeanu}{Cyclic Neural-Analytic Approach (CNA)}

\usepackage{pifont}
\usepackage{xcolor}
\usepackage{graphicx}

\definecolor{m_green}{HTML}{2C8915}
\definecolor{m_red}{HTML}{FF0000}

\usepackage{diagbox}

\begin{document}

\maketitle

\begin{abstract}
Generating novel views from recorded videos is crucial for enabling autonomous UAV navigation. Recent advancements in neural rendering have facilitated the rapid development of methods capable of rendering new trajectories. However, these methods often fail to generalize well to regions far from the training data without an optimized flight path, leading to suboptimal reconstructions. We propose a self-supervised cyclic neural-analytic pipeline that combines high-quality neural rendering outputs with precise geometric insights from analytical methods. %Our solution enhances both RGB and mesh reconstructions for novel view synthesis, particularly in undersampled areas and regions entirely distinct from the training dataset. 
Our solution improves RGB and mesh reconstructions for novel view synthesis, especially in undersampled areas and regions that are completely different from the training dataset.
We use an effective transformer-based architecture for image reconstruction to refine and adapt the synthesis process, enabling effective handling of novel, unseen poses without relying on extensive labeled datasets. Our findings demonstrate substantial improvements in rendering views of novel and also 3D reconstruction, which to the best of our knowledge is a first, setting a new standard for autonomous navigation in complex outdoor environments. To foster further research and application, we will make our code publicly available.
\end{abstract}

%-------------------------------------------------------------------------

\section{Introduction}
\label{sec:intro}

While image and video generative AI models excel at crafting novel scenes from brief textual descriptions, their utility extends beyond mere visual creativity to critical real-world applications. One such pivotal application is the simulation of virtual flights over actual landscapes, which plays a crucial role in autonomous navigation and the development of digital twins. Traditional approaches to novel view synthesis predominantly focus on object-centered or synthetic datasets characterized by minimal noise in the input data. 
%Testing typically occurs on the same images used for training or at regular intervals within a sequence, such as every \textit{k} frame \cite{barron2022mip}. 
Testing typically uses the same images as training or occurs at regular intervals within a sequence, such as every k-th frame\cite{barron2022mip}.
Although current neural rendering techniques can produce high-quality reconstructions on the training data -- with PSNR values often exceeding 30 \cite{kerbl3Dgaussians}. Due to the ray-based nature used to understand the world, camera paths that are not close to the training data result in very poor reconstructions. As a result, camera paths diverging significantly from the training data frequently lead to drastically inferior reconstructions. A prevalent strategy to mitigate these limitations involves segmenting datasets into smaller, consistent regions \cite{meuleman2023localrf, tancik2022block}. However, this approach demands substantial computational resources, as a separate model must be trained for each sequence \cite{li2024nerfxl, suzuki2024fed3dgs}. This not only increases the complexity but also limits the scalability of such solutions, underscoring the need for more robust, generalized methods capable of handling diverse and dynamic environments.

\vspace{1mm}
\noindent We introduce a pioneering cyclic neural-analytic approach (CNA) that synergistically utilizes the strengths of structure-from-motion and neural rendering methods to synthesize high-fidelity RGB images on novel poses far away from what it saw during training. Central to our approach is the integration of a low computational footprint with advanced self-supervised learning techniques, facilitated by our transformer-based neural branch. This enables effective synthesis even from a sparse dataset of approximately 400 images, completing the video-to-mesh transformation within about an hour, contingent on the scene's geometric complexity, as we show in our experiments. 

\vspace{1mm}
\noindent Our pipeline significantly enhances this process by employing a dual-phase reconstruction strategy. Initially, an analytic 3D reconstruction aligns with the neural rendering branch outputs to refine the geometric consistency. Subsequently, we repurpose a lightweight transformer for image restoration, to achieve high-resolution reconstruction of novel 2D views. Unique to our system is the rapid, scene-specific fine-tuning capability of this self-supervised module, allowing it to adapt seamlessly to diverse scenarios independently—contrast this with traditional, larger models that typically require extensive datasets and substantial computational resources. 

\vspace{1mm}
\noindent By synergistically leveraging both analytic insights and neural rendering, our method not only reduces computational demands but also maximizes the efficiency of the transformer architecture, enriched with a novel reconstruction prior. Tailored to individual scenes, it adeptly captures unique scene characteristics without necessitating recalibration or extensive retraining. The outcome is a robust system capable of delivering photorealistic reconstructions from a minimal dataset, providing a scalable solution ideally suited for applications in virtual reality, augmented reality, and robotic vision.

\vspace{1mm}
\noindent Our \textbf{main contributions} are summarized below:
\vspace{-2mm}
\begin{enumerate}
\item We combine an analytical and neural 3D reconstruction method through a novel cyclic self-supervised transformer-based approach and show significant improvements for both novel view synthesis, as well as 3D reconstruction through iterative learning.  
\vspace{-2mm}
\item Our framework demonstrates robust generalization capabilities, effectively generating high-quality images from test locations without the need for additional training data.
\vspace{-2mm}
\item We show significant improvement compared to state-of-the-art methods on very difficult cases of real-world scenes, captured by UAVs, that spawn a large area, not just object-centric, having a significant amount of noise in pose and depth. 
\vspace{-4mm}
\end{enumerate}

%-------------------------------------------------------------------------

\begin{figure*}[!ht]
\begin{center}
\includegraphics[width=\linewidth]{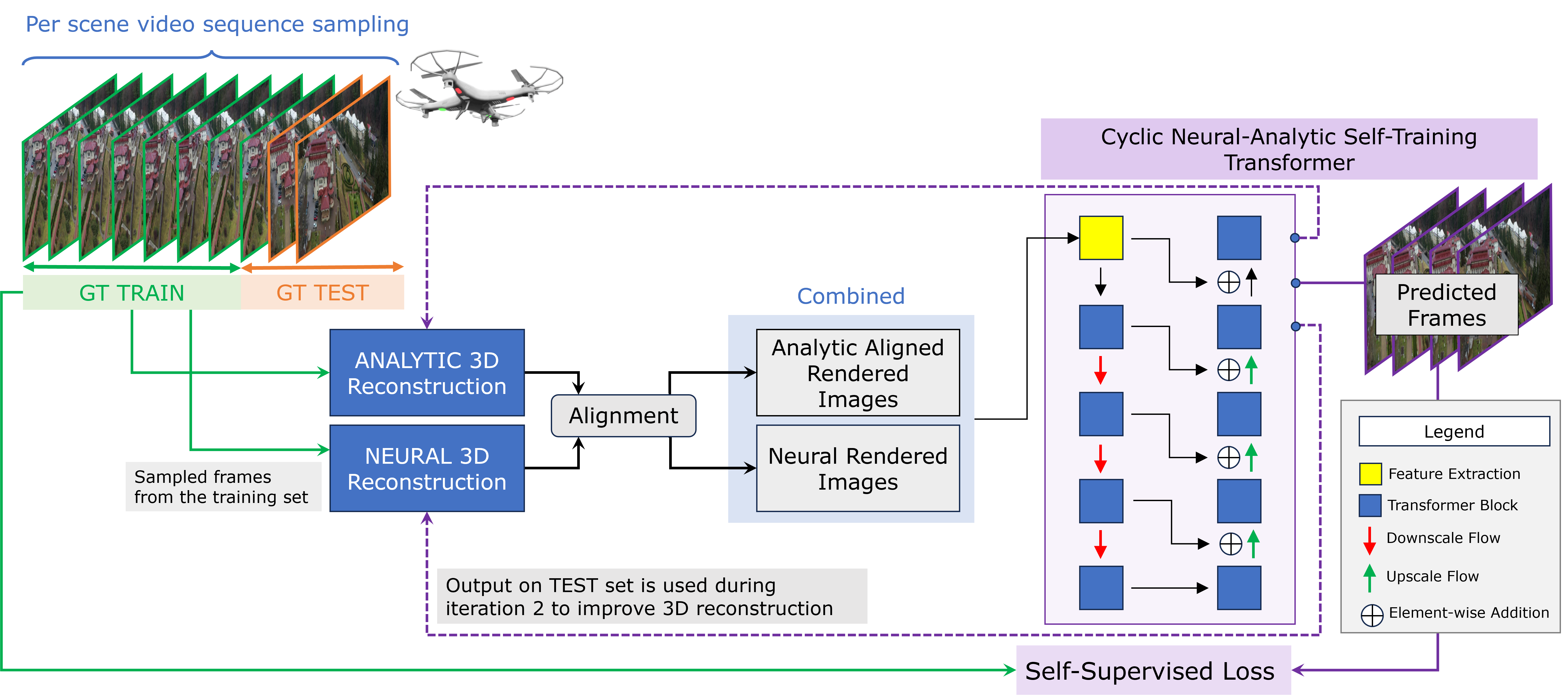}
\vspace{-7mm}
\caption{An overview of our novel self-supervised cyclic neural-analytic pipeline for novel 2D view synthesis. We rely on both traditional and modern 3D reconstruction methods which we combine through a self-supervised transformer-based U-net style model for improved image reconstruction. We employ an iterative learning procedure in which the outputs from the first learning iteration become inputs for the next to further refine the results in terms of RGB and mesh, without additional new images. We work in the UAV video domain and use the last 20\% from the image sequence as testing to simulate a more realistic reconstruction scenario. \textbf{Original RGB frames from the TEST set are used exclusively for evaluation purposes.}}\label{fig:overview}
\end{center}
\vspace{-10mm}
\end{figure*}

%-------------------------------------------------------------------------

\section{Related Work}
\label{sec:related_work}

\textbf{Analytic 3D reconstruction} -- Classic feature-based structure-from-motion methods such as COLMAP~\cite{schoenberger2016sfm,griwodz2021alicevision} require significant processing time, even for a small number of high-resolution images~\cite{jiang2020efficient}. As the number of images increases, the resources required increase exponentially~\cite{triggs2000bundle}. 

\noindent\textbf{Neural 3D reconstruction} -- The implicit representation proposed by Neural Radiance Fields (NeRF)~\cite{mildenhall2021nerf, zhang2020nerf++}, with faster versions such as Instant Neural Graphics Primitives~\cite{muller2022instant}, NerfAcc~\cite{li2022nerfacc} and NeuS~\cite{wang2021neus, neus2}, drastically reduced the computation time to several minutes per scene. %Different from them, ours is constrained by the initial 3D reconstruction prior, which makes it less prone to overfitting, as our experiments will show. Direct Voxel Grid Optimization~\cite{sun2022direct} proposes a representation that replaces the MLP usually used in NeRFs with a voxel grid, resulting in speed and performance gains. 
Unlike these methods, our approach is constrained by the initial 3D reconstruction prior, which makes it less prone to overfitting, as our experiments will show. Direct Voxel Grid Optimization~\cite{sun2022direct} proposes a representation that replaces the MLP usually used in NeRFs with a voxel grid, resulting in speed and performance gains. 
Another downside of recent works is the need for accurate pose information. Neural bundle adjustment methods were developed to address this issue~\cite{lin2021barf, chen2022local, wang2023posediffusion, lin2023relposepp}, but this means another step and they generally work on a centered object.

\noindent\textbf{Scene reconstruction} -- Given the errors that occur over a large set of images, some authors proposed splitting the scene into several smaller ones on a grid~\cite{tancik2022block,turki2023suds,xu2023gridguided}, splitting the path into image sets~\cite{meuleman2023localrf}, or splitting the scene into smaller NeRFs~\cite{turki2022mega, zhang2023nerflets, cheng2023lu}. These methods are not globally consistent and hide the pose and reconstruction errors. Furthermore, they usually require significant resources and training.

\noindent\textbf{Mesh-based reconstruction} -- Our method benefits from the rendered mesh created in the analytical phase. While some methods directly output textured mesh such as~\cite{tang2022nerf2mesh}, they are generally limited to small objects. A recent method that combines mesh constraints and neural rendering is~\cite{guedon2023sugar}, but they explicitly generate the mesh, as opposed to our method, which requires no explicit mesh understanding.

\noindent\textbf{Our work in context} -- A key contribution of our work is the ability to outperform state-of-the-art methods from views that are drastically different from the training cases, using only a small sequence of RGB images, even in the challenging case of faraway objects. Furthermore, our pipeline is method agnostic - any two neural and analytical methods can be combined to generate our cyclic method. 

%-------------------------------------------------------------------------

\section{Self-supervised Cyclic Neural-Analytic Approach}
\label{sec:method}

Our pipeline consists of two complementary modules - an analytical and a neural one. Fundamentally, both are capable of rendering images from novel poses. The initial iteration uses the training set for both modules. New images are generated from the test poses and included in the training data. We repeat the cycle with the newly generated images. No information is shared between the two apart from the input image set and the index of the pose that needs to be rendered. A visual overview of our pipeline can be depicted in Figure~\ref{fig:overview} and we offer more details about each component from our self-supervised cyclic neural-analytic pipeline below.

\subsection{Analytic 3D Reconstruction}

Traditional 3D reconstruction methods reconstruct three-dimensional structures from two-dimensional image sequences, estimating both the 3D structure of a scene and the cameras' poses. We employ a standard photogrammetric 3D reconstruction pipeline, such as~\cite{schoenberger2016sfm, schoenberger2016mvs, griwodz2021alicevision}. The pipeline processes a set of unordered images to reconstruct 3D scenes, through image-based feature extraction using SIFT and AKAZE and then matching these across images to establish correspondences. Following this, the structure-from-motion (SfM) phase computes the 3D positions of these features and the camera poses, creating a sparse 3D model. This model is then densified through Multi-View Stereo (MVS), which refines it into a dense point cloud. The cloud is transformed into a mesh, optimized for smoothness and continuity, and finally textured using the best views from the input images providing both geometric detail and realistic texturing. Besides the computationally intensive and time-consuming nature of the current pipeline, other limitations are linked to its dependency on feature detection and matching. Specifically, 1) SfM struggles in low-texture or repetitively patterned environments where detecting distinct features becomes problematic, and 2) dynamic changes or obstructions in the scene can compromise the completeness and accuracy of the reconstructions. Leveraging neural rendering techniques that do not rely strictly on feature matches but instead use a continuous representation of the scene, which can robustly handle texture-poor environments and dynamic changes, enhances both the efficiency and fidelity of the reconstruction process.

\subsection{Neural 3D Reconstruction}

While there are many neural methods in the NeRF family for scene reconstruction, we leverage the  Gaussian Splatting~\cite{kerbl3Dgaussians} method. This method introduces a highly efficient approach for rendering radiance fields in real-time by employing a novel scene representation using three-dimensional Gaussians. Unlike traditional methods that rely on dense neural networks and volumetric ray-marching, this approach uses sparsely distributed 3D Gaussians derived from camera calibration points obtained through SfM. These Gaussians efficiently capture the significant geometric features of the scene while minimizing computations in empty space. The scene is represented using 3D Gaussians, each characterized by its position, an anisotropic covariance matrix that defines its shape and orientation, and an opacity value. The Gaussian \( G(x) \) at a position \( x \) in the scene is mathematically defined as follows:
\[
G(x) = \alpha \exp\left(-\frac{1}{2}(x - \mu)^T \Sigma^{-1} (x - \mu)\right)
\]
where \( \mu \) is the mean or the center position of the Gaussian, \( \Sigma \) is the covariance matrix determining the spread and orientation, and \( \alpha \) denotes the opacity. This representation efficiently captures the scene's essential details without dense sampling. The properties of each Gaussian, such as position, shape, and opacity, are optimized iteratively. This optimization adjusts the density and distribution of the Gaussians based on the rendering needs, striking a balance between detail fidelity and computational efficiency. The rendering process utilizes a tile-based rasterization approach, projecting the 3D Gaussians onto a 2D plane for efficient blending. The method maintains the correct visibility order through depth-based sorting of the Gaussians, crucial for preserving image quality through proper alpha compositing. Combining the representational efficiency of 3D Gaussians with the rapid processing of tile-based rasterization, the method achieves real-time rendering capabilities. It significantly surpasses traditional NeRF implementations in training speed and rendering time, making it viable for high-resolution, real-time applications.  

\subsection{Self-supervised Neural-Analytic Novel View Synthesis}

We combine the previous analytical and neural 3D reconstruction by adapting a lightweight transformer-based architecture~\cite{zhao2023comprehensive} for enhanced image reconstruction. While our training pipeline is compatible with various model types, we chose not to compromise on quick inference speeds and also aimed for satisfactory performance, which this lightweight model successfully delivered. The model highly resembles a U-net~\cite{ronneberger2015u} style architecture. It leverages a hierarchical multi-scale approach based on an encoder-decoder configuration to efficiently handle image restoration tasks. The architecture is depicted in Figure~\ref{fig:overview}, the encoder progressively downscales the input features, while it increases the number of feature maps by a factor of 2, at every scale. The decoder progressively upscales the outputs from each transformer block from the encoder branch after an element-wise addition between output from the encoder and the projected output from the decoder.  

\noindent The main innovation of this model relies on the transformer blocks used instead of simple convolutional layers compared to the original. The architecture uses two primary neural blocks - the condensed attention neural block and the dual adaptive neural block, which operate sequentially at each scale of the encoder and decoder. Detailing these blocks is beyond the scope of this submission and we refer the reader to the original paper.

We first align the analytical reconstruction maps to the neural reconstruction maps and we concatenate these results along the last dimension (channel-wise). We train the model in a self-supervised manner by using RGB frames from video sequences as sole supervision ($R_{ijk}$) when computing the pixel-wise mean squared error loss on predictions ($P_{ijk}$): 

\[
Loss = \frac{1}{N} \sum_{i=1}^H \sum_{j=1}^W \sum_{k=1}^C (P_{ijk} - R_{ijk})^2
\]

\text{Where:}
\begin{itemize}
    \item \( P_{ijk} \) and \( R_{ijk} \) are the values at position \( (i, j, k) \) in the prediction and reference maps, respectively.
    \item \( H, W, \) and \( C \) represent the height, width, and number of channels in the maps.
    \item \( N = H \times W \times C \) is the total number of elements (including all channels) in each map.
\end{itemize}
%More details about the training procedure are provided in the supplementary material. 
% DONE - add the details from the supplementary material here
\noindent\textbf{Model Architecture.} We used CODEFormer~\cite{zhao2023comprehensive} to combine the neural and analytic branches within our Cyclic Neural-Analytic (CNA) approach. CODEFormer is a transformer-based architecture designed for image restoration tasks such as Gaussian denoising, JPEG compression artifact reduction, and mixed degradation. The model consists of an encoder-decoder structure with additional refinement blocks for enhanced performance. Key components of the model include a patch embedding layer that converts input images into a sequence of patches, an encoder with a series of transformer blocks that progressively downsample the input features, a decoder with a corresponding series of transformer blocks that upsample the features back to the original resolution, refinement blocks that further process the features before generating the final output, and convolutional layers used in patch embedding and final reconstruction. The model's hyperparameters and structural details include input channels set to 6, embedding dimension of 48, number of blocks [4, 6, 6, 8], channel squeeze ratios [4, 2, 2, 1], number of shuffles [1, 2, 4, 8], expanded attention channels of 16, and 4 refinement blocks.

\noindent\textbf{Training Setup.} Experiments were conducted using the same set of parameters, which include a batch size of 2 and patch size of 128. The optimizer used was Adam with an initial learning rate of 0.0001 with a cosine annealing learning rate scheduler for a total of 200 epochs. We did not conduct any hyperparameters optimization per dataset. In terms of training infrastructure, all our experiments were conducted using one GeForce RTX 4090 Graphics Cards. 

\vspace{-2mm}
\subsection{3D Cyclic Refinement}

Our method includes a cyclic refinement that improves upon the results of the first iteration without any additional supervision. Only by feeding the rendered images from the test poses to the algorithm do we notice a performance gain in terms of image reconstruction.  

%-------------------------------------------------------------------------
\vspace{-2mm}
\section{Experimental Analysis}
\label{sec:experiments}

We apply our cyclic neural-analytic framework to a wide range of outdoor scenes. We aim to deliver improved performance on the testing set without any additional labels or retraining. Our experimental analysis demonstrates consistent improvements over iterations. Although our primary focus is on large-scale outdoor scenes, we are also interested in contrasting our results with those from an object-centric dataset, which typically exhibits robust performance when using neural methods.

\noindent\textbf{Aerial dataset} -- Aerial scene reconstruction, as opposed to a single-centered object generally targeted by fast reconstruction methods. We use the public real-world Aerial dataset proposed in~\cite{licuaret2022ufo}, which features telemetry data (commonly provided by UAV manufacturers) and a diverse set of landscapes, with both vegetation and man-made structures. The flight altitude is generally 50m and the flight trajectory is manual. Although the original resolution is 4K (3180 × 2160 px), we rescale them to 1920 × 1080 for most experiments. We use 5-minute videos (9000 frames in total) which we sample every 20 frames. We report results for all four scenes from the dataset - Slanic, Olanesti, Chilia, and Herculane. 
\vspace{1mm}

\noindent\textbf{BlendedMVS} -- We have selected an outdoor scene reconstruction from the BlendedMVS~\cite{yao2020blendedmvs} dataset. While it is lower resolution compared to the Aerial dataset, it consists of a similar outdoor scenario and the images originate from a rendered reconstructed mesh. As opposed to the other datasets that generally feature a clip with continuous frames, this scenario involves a grid of points (check supplementary material for details). 
\vspace{1mm}

\noindent\textbf{Rubble} -- We have selected a drone-captured dataset with plenty of details - Mill 19 - Rubble~\cite{Turki_2022_CVPR} to compare how well our method can improve upon unstructured details. Feature matching is more prone to false positives and while images are high resolution (4608 × 3456), details are difficult to resolve.
\vspace{1mm}

\noindent\textbf{Tanks and Temples} -- We use the Family scene from Tanks and Temples~\cite{Knapitsch2017} to evaluate on the object-centric scenario. It consists of a ground-level video captured around a statue. The original clip was sampled every 10 frames, we resampled from the original clip for a total of 440 frames.
\vspace{1mm}

\noindent We split each scene of the dataset into two parts: the reconstruction part (train set) and the reprojection part (test set). The reconstruction part contains 80\% of the data points, leaving the rest of 20\% as the reprojection data. We reconstruct the scene based on only the reconstruction data and test the results using the reprojection data. In this way, we ensure that the results are based on data that was not used during the reconstruction process. During the reprojection phase, we use the poses of the drone to place a virtual camera with the same intrinsic parameters as the intrinsics of the drone and capture an RGB image for each pose.

\subsection{Comparison to state-of-the-art}
\label{sec:comparisons}

We compare our method against both implicit and explicit depth methods with similar computing requirements. Instant-NGP~\cite{muller2022instant} is a fast framework that benefits from a 2D location hashing scheme~\cite{tinycudann} and a CUDA-optimized MLP architecture. We compare two versions - vanilla and depth input. The depth support was added after the initial code release and was not officially supported. Neuralangelo~\cite{li2023neuralangelo} is a significant improvement over Instant-NGP in terms of reconstruction quality, but despite following the authors' guidelines and higher compute requirements (minutes vs. a whole day, the resulting model has $\approx$300M parameters), we found it would only overfit better on the training data in our scenario.

\noindent We have also selected two complementary methods that aim to better model the depth. The first one, Direct Voxel Grid Optimization~\cite{sun2022direct} is similar to a NeRF that has the neural network replaced by a dense voxel grid. On the other hand, Plenoxels~\cite{yu_and_fridovichkeil2021plenoxels} uses a sparse grid to build a more efficient scene representation.

\begin{figure*}[t!]
\begin{center}
\includegraphics[width=0.8\textwidth]{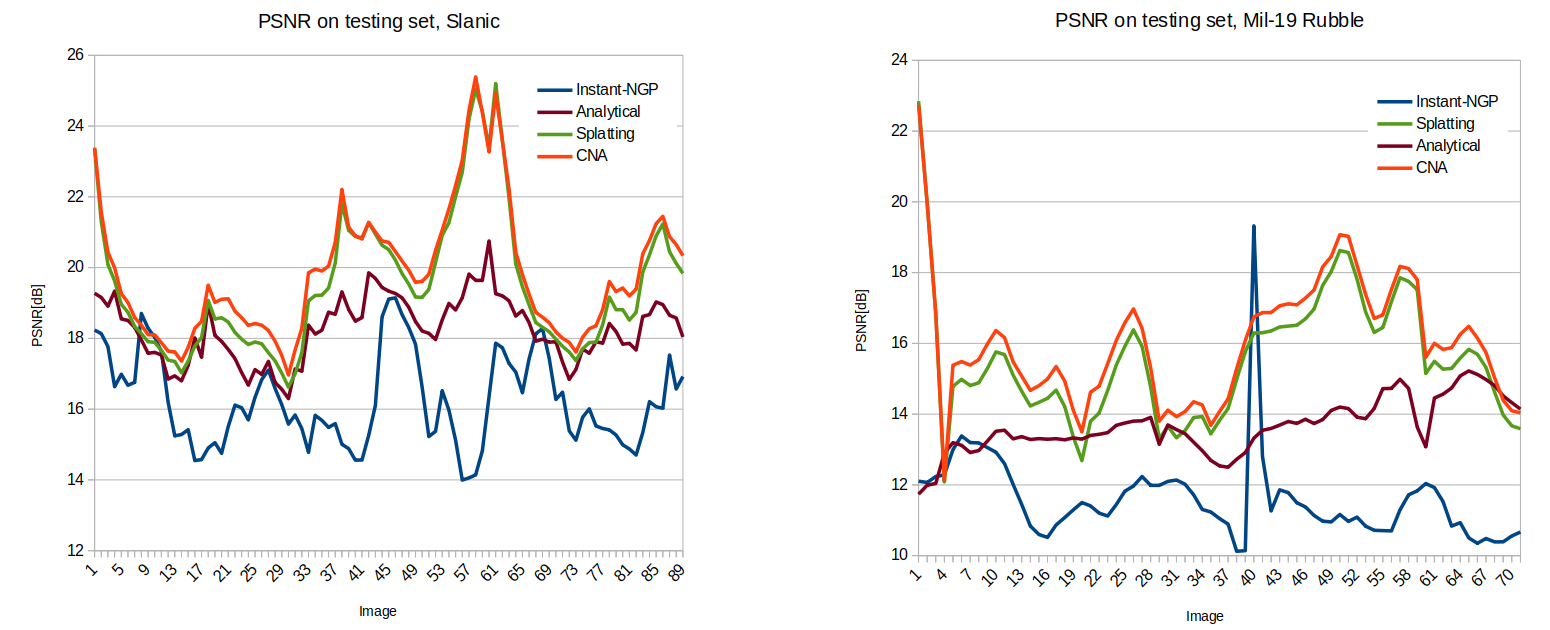}
\vspace{-2mm}
\caption{PSNR error on Slanic on each test frame. CNA consistently improves over the other baselines, despite not receiving additional RGB images. The performance gains are not only average, but on the vast majority of frames. Full results are shown in the supplementary material.}\label{fig:psnr_error}
\end{center}
\vspace{-8mm}
\end{figure*}

\noindent Finally, we compare with Gaussian Splatting~\cite{kerbl3Dgaussians} - a real-time rendering method that shows impressive performance on real scenes. Nevertheless, as the camera diverges from the training trajectory, performance drops. The results on the Aerial dataset are shown in Table \ref{tab:results_reconstruction_scene}. Since we focus on UAV scene reconstruction, we provide detailed comparisons of all scenes. CNA yields an improved result without any additional labels and most of the time the improvements are consistent across scenes. The results on other datasets are shown in Table \ref{tab:results_sota_other_datasets}. We note an improvement on all types of scenes, regardless if it is object-centric (Tanks and Temples), drone-captured (Rubble), or synthetically reconstructed (BlendedMVS). The object-centric scenario is particularly difficult to improve upon as most neural methods are designed for this task. As shown in Figure \ref{fig:psnr_error}, the performance gains are not only on average but on most of the frames.

\subsection{Ablation studies}

\begin{figure}[!ht]
\begin{center}
\includegraphics[width=0.84\textwidth]{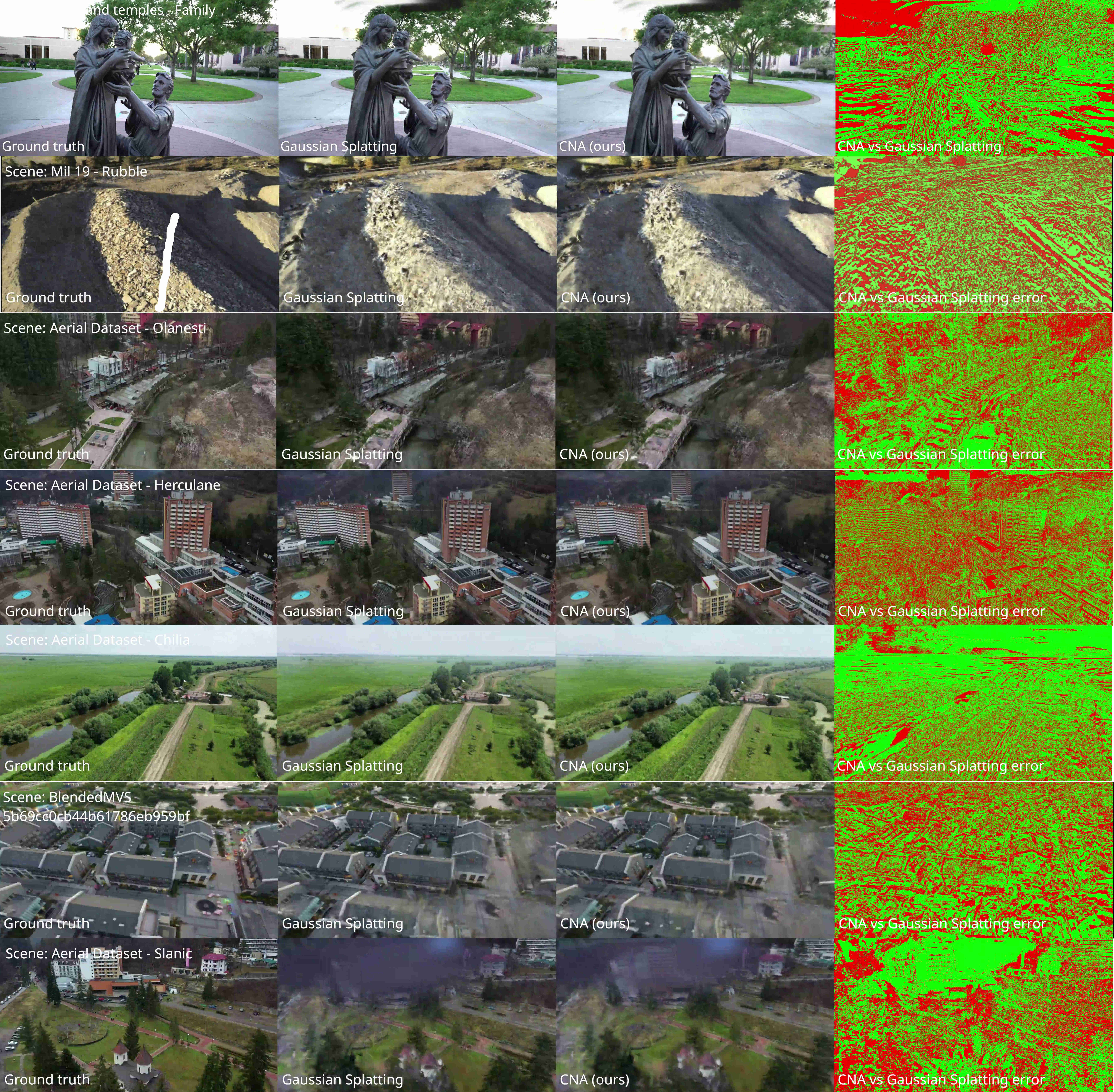}
\vspace{-2mm}
\caption{Qualitative results. From left to right, RGB, Gaussian Splatting, CNA, CNA vs Gaussian Splatting error. Green means CNA has a smaller error. The errors from Gaussian Splatting are harder to spot due to the blending, but poorly seen regions and artifacts are better dealt with using CNA.}\label{fig:qual}
\end{center}
\vspace{-6mm}
\end{figure}

We conduct an ablation study to show the improvement across iterations on the Aerial dataset and the contribution of each component for our pipeline -- analytic and neural. As shown in Table~\ref{tab:ablation_studies} in blue, our method consistently yields better results, without additional labels. Most results from iteration 2 outperform iteration 1, regardless of the method they are applied to. Even though the analytical method yields poorer-quality images compared to the neural one, alignment clearly helps and iterations help boost the final score. Note that further gains may be obtained by unifying poses for the two methods -- as it stands now, a separate estimation is run for each method. We show qualitative results in Figure~\ref{fig:qual}. We compared the closest competitor, Gaussian Splatting with CNA. Green means our method has a smaller error compared to ground truth. Our method yields an improvement, especially in the regions where artifacts from Gaussian Splatting are present. Nevertheless, the analytical grounding also helps in other problematic regions, such as occlusions (see last row). 

\begingroup
\setlength{\tabcolsep}{4pt}
\begin{table*}[t]
\begin{center}
\caption{\label{tab:ablation_studies}
Comparison to state-of-the-art methods for novel view synthesis. PSNR reconstruction results on the Aerial dataset~\cite{licuaret2022ufo} for the test set exclusively. The best numbers for each set are \textbf{bolded}.}
\vspace{2mm}
\begin{tabular}{|l|c|c|c|c|c|}
\hline
\backslashbox{Method}{Scene}
 & Slanic & Olanesti & Chilia & Herculane & Mean\\ \hline\hline

SFM\hfill~\cite{griwodz2021alicevision} & 18.43 & 17.63 & 18.75 & 18.27 & 18.27 \\ \hline

DVGO\hfill~\cite{sun2022direct} & 15.34 & 14.34 & 16.43 & 15.43 & 15.38\\ \hline

Plenoxels\hfill~\cite{yu_and_fridovichkeil2021plenoxels} & 17.22 & 15.92 & 18.59 & 16.07 & 15.38\\ \hline

Instant-NGP\hfill~\cite{muller2022instant} & 16.54 & 19.25 & 21.01 & 19.12 & 18.98 \\ \hline

Neuralangelo\hfill~\cite{li2023neuralangelo} & 16.79 & 16.20 & 16.96 & 15.82 & 17.29 \\ \hline

GeoView\hfill~\cite{budisteanu2023selfsupervised} & 20.28 & 18.64 & 19.04 & 18.17 & 19.03 \\ \hline

Gaussian Splatting\hfill~\cite{kerbl20233d} & 19.37 & 19.34 & 21.85 & 20.85 & 20.35 \\ \hline \hline

CNA (Ours) & \textbf{19.85} & \textbf{19.81} & \textbf{22.21} & \textbf{21.39} & \textbf{20.82} \\ \hline
\end{tabular}
\end{center}
\vspace{-3mm}
\end{table*}
\endgroup

\begingroup
\setlength{\tabcolsep}{4pt}
\begin{table*}[t]
\begin{center}
\caption{\label{tab:results_sota_other_datasets}
Test set comparison with Instant-NGP, Gaussian Splatting and the baseline analytical reconstruction method on three additional scenes. We have selected scene \textit{5b69cc0cb44b61786eb959bf} from BlendedMVS, the Rubble scene from MIL-19, and the Family scene for Tanks and Temples. The split structure is the same as with our previous experiments (80\% training, 20\% testing). The best numbers for each set are \textbf{bolded}.}
\vspace{2mm}
\begin{tabular}{|l|c|c|c|c|}
\hline
\backslashbox{Method}{Scene}
 & BlendedMVS & Rubble & Tanks and temples& Mean\\ \hline

Instant-NGP~\cite{muller2022instant} & 14.54 & 11.54 & 14.61 & 13.56 \\ \hline

Analytical~\cite{griwodz2021alicevision} & 13.95 & 13.65 & 15.73 & 14.44\\ \hline

Gauss.Splatt.~\cite{kerbl20233d} & 21.52 & 15.53 & 19.63 & 18.89  \\ \hline

CNA (Ours) & \textbf{21.67} & \textbf{16.04} & \textbf{19.81} & \textbf{19.17} \\ \hline
\end{tabular}
\end{center}
\vspace{-5mm}
\end{table*}
\endgroup

\begingroup
\setlength{\tabcolsep}{2pt}
\begin{table*}[t]
\begin{center}
\caption{\label{tab:results_reconstruction_scene}
Ablative studies for each of the components in our proposed method, CNA on the Aerial dataset~\cite{licuaret2022ufo}. PSNR reconstruction results are reported on the test set exclusively. (1) and (2) denote the first and second iteration of CNA. The best are  \textbf{bolded}. Blue values denote iteration-level improvement (the best values for each).}
\vspace{2mm}
\begin{tabular}{|l|c|c|c|c|c|c|c|c|c|c|}
\hline
\backslashbox{Method}{Scene}
 & \multicolumn{2}{|c|}{ Slanic } & \multicolumn{2}{|c|}{ Olanesti } & \multicolumn{2}{|c|}{ Chilia } & \multicolumn{2}{|c|}{ Herculane }  & \multicolumn{2}{|c|}{ Mean } \\ \hline\hline
 %& It. (1) & It. (2) & It. (1) & It. (2) & It. (1) & It. (2) & It. (1) & It. (2) & It. (1) & It. (2)\\ \hline
Iteration & (1) & (2) & (1) & (2) & (1) & (2) & (1) & (2) & (1) & (2)\\ \hline

Analytic (only) & 18.43 & \color{blue}{18.66} & \color{blue}{17.63} & 16.99 & 18.75 & \color{blue}{19.12} & 18.27 & \color{blue}{18.83} & 18.27 & \color{blue}{18.40}\\ \hline

Analytic (aligned) & 19.11 & \color{blue}{19.33} & \color{blue}{17.73} & 17.05 & 18.95 & \color{blue}{19.81} & 18.42 & \color{blue}{19.01} & 18.55 & \color{blue}{18.80}\\ \hline

Neural (only) & 19.49 & \color{blue}{19.73} & 19.48 & \color{blue}{\textbf{19.85}} & 21.28 & \color{blue}{22.19} & 21.27 & \color{blue}{21.38} & 20.38 & \color{blue}{20.79}\\ \hline \hline

CNA (Ours) & 19.65 & \color{blue}{\textbf{19.85}} & \color{blue}{19.81} & \color{blue}{19.81} & 21.90 & \color{blue}{\textbf{22.21}} & 21.32 & \color{blue}{\textbf{21.39}} & 20.67 & \color{blue}{\textbf{20.82}}\\ \hline
%FlightDreamer (with TripleHop) & TODO & TODO & - & - & -  & - & - & - & -  & - \\ \hline
\end{tabular}
\end{center}
\vspace{-8mm}
\end{table*}
\endgroup

\vspace{-2mm}
\subsection{Improving 3D Reconstruction}

\begin{figure*}[!ht]
\begin{center}
\includegraphics[width=\linewidth]{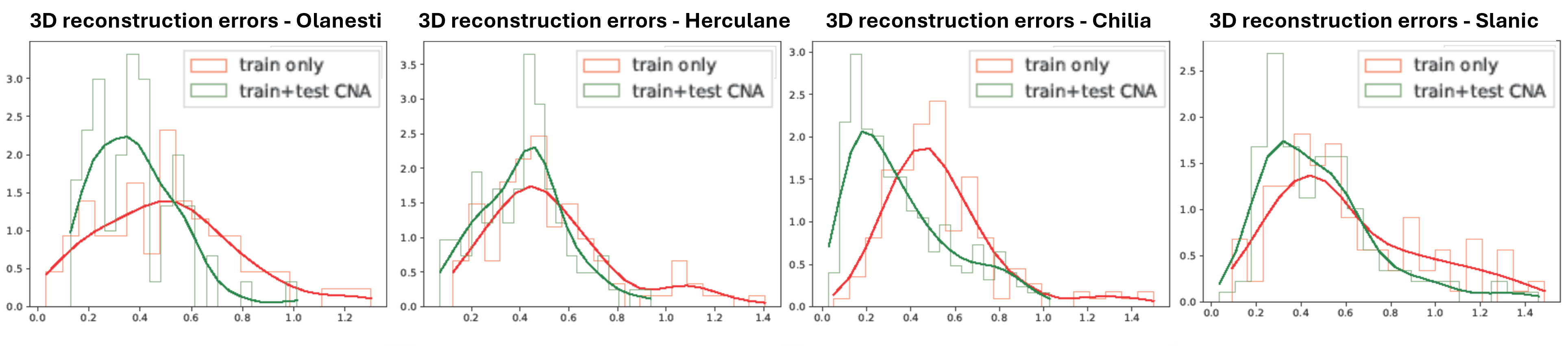}
\vspace{-8mm}
\caption{Mesh reconstruction improvement evaluation for different scenes - training vs training and generated set. We first obtain a ground truth mesh from the combined training an testing set. We align and compare it with (a) the one featuring only training images and (b) training images with generated test images (no additional information). A histogram over the error bins is shown. Our method (green) consistently improves over the mesh generated with training images only(red), significantly reducing the large 3D reconstruction errors.}
\label{fig:improving_3d}
\end{center}
\vspace{-10mm}
\end{figure*}

We assess the quality of the reconstructed mesh using the one generated from training and testing combined. We compare the training-only mesh with the one enhanced by the CNA-rendered set. To do this, we align two meshes to a reference mesh. The reference mesh is the one built using the whole dataset. The second mesh is built with training images and training images coupled with CNA-rendered images. Results are shown as histograms in Figure \ref{fig:improving_3d} and Table \ref{tab:results_mesh_reconstruction}.

\noindent\textbf{Impact and limitations} -- Novel view synthesis with explicit depth representation is a helpful tool for safer navigation or environmental monitoring. With the ever-increasing resolution capabilities of UAV cameras, our pipeline could improve the performance of modern neural reconstruction pipelines. Nevertheless, the iterative part adds a compute overhead and we are looking into ways of making the pipeline faster. A simple solution would be to use only the neural part but that would miss on additional performance gains. 

\begingroup
\setlength{\tabcolsep}{3pt}
\begin{table*}[!ht]
\begin{center}
\vspace{1mm}
\caption{\label{tab:results_mesh_reconstruction}
3D reconstruction errors in \textbf{meters} over iterations on the Aerial dataset~\cite{licuaret2022ufo} for the test set exclusively. Best numbers for each individual set are \textbf{bolded}. Since the depth range is 30-300m, this translates into an $\approx$1\% error.}
%\vspace{5mm} % wth
\begin{tabular}{|l|c|c|c|c|c|c|c|c|}
\hline
\backslashbox{Method}{Scene}
 & \multicolumn{2}{|c|}{ Slanic } & \multicolumn{2}{|c|}{ Olanesti } & \multicolumn{2}{|c|}{ Chilia } & \multicolumn{2}{|c|}{ Herculane } \\ \hline\hline
 & Mean & Median & Mean & Median & Mean & Median & Mean & Median \\ \hline
 
Iteration 1 & 0.6102 & 0.5263 & 0.5053 & 0.5013 & 0.5324 & 0.4932 & 0.5289 & 0.4802 \\ \hline
Iteration 2 & \textbf{0.4832} & \textbf{0.4392} & \textbf{0.3806} & \textbf{0.3646} & \textbf{0.3553} & \textbf{0.2934} & \textbf{0.4125} & \textbf{0.4225} \\ \hline

\end{tabular}
\end{center}
\vspace{-8mm}
\end{table*}
\endgroup

\section{Conclusions}
\label{sec:conclusions}

We present a self-supervised cyclic neural-analytic pipeline that blends neural rendering with mesh-based analytical methods. Our solution enhances both RGB and 3D mesh reconstructions for novel view poses that are significantly different from the training set.
Extensive experiments on a wide range of datasets demonstrated the effectiveness of both the neural and analytic modules and the usefulness of cycling. Our design that allows drop-in replacement of modules covers new ground in this area and has the potential to push the boundaries further towards novel view synthesis that is grounded in the 3D physical world.

\vspace{-3mm}

\vspace{2mm}
% New
\noindent\textbf{Acknowledgements} -- This work was supported in part by the EU Horizon project ELIAS (EU Horizon project ELIAS (Grant number: 101120237) and project "Next Generation 3D Machine Vision with Embedded Visual Computing" funded by Research Council of Norway (Grant number: 325748). We also want to express our sincere gratitude to Aurelian Marcu from NILPRP for providing access to GPU resources from the National Interest Infrastructure facility IOSIN-CETAL.

% Old 
%\noindent\textbf{Acknowledgements} -- This work was supported by the EU Horizon project ELIAS (No. 101120237). We also want to express our sincere gratitude to Aurelian Marcu from NILPRP for providing access to GPU resources from the National Interest Infrastructure facility IOSIN-CETAL.
\vspace{-2mm}

\bibliography{egbib}

\begin{thebibliography}{40}
\providecommand{\natexlab}[1]{#1}
\providecommand{\url}[1]{\texttt{#1}}
\expandafter\ifx\csname urlstyle\endcsname\relax
  \providecommand{\doi}[1]{doi: #1}\else
  \providecommand{\doi}{doi: \begingroup \urlstyle{rm}\Url}\fi

\bibitem[Barron et~al.(2022)Barron, Mildenhall, Verbin, Srinivasan, and Hedman]{barron2022mip}
Jonathan~T Barron, Ben Mildenhall, Dor Verbin, Pratul~P Srinivasan, and Peter Hedman.
\newblock Mip-nerf 360: Unbounded anti-aliased neural radiance fields.
\newblock In \emph{Proceedings of the IEEE/CVF Conference on Computer Vision and Pattern Recognition}, pages 5470--5479, 2022.

\bibitem[Budisteanu et~al.(2023)Budisteanu, Costea, Marcu, and Leordeanu]{budisteanu2023selfsupervised}
Alexandra Budisteanu, Dragos Costea, Alina Marcu, and Marius Leordeanu.
\newblock Self-supervised novel 2d view synthesis of large-scale scenes with efficient multi-scale voxel carving, 2023.

\bibitem[Chen et~al.(2022)Chen, Chen, Wang, Zhang, Guo, Shan, and Wang]{chen2022local}
Yue Chen, Xingyu Chen, Xuan Wang, Qi~Zhang, Yu~Guo, Ying Shan, and Fei Wang.
\newblock Local-to-global registration for bundle-adjusting neural radiance fields.
\newblock \emph{arXiv preprint arXiv:2211.11505}, 2022.

\bibitem[Cheng et~al.(2023)Cheng, Esteves, Jampani, Kar, Maji, and Makadia]{cheng2023lu}
Zezhou Cheng, Carlos Esteves, Varun Jampani, Abhishek Kar, Subhransu Maji, and Ameesh Makadia.
\newblock Lu-nerf: Scene and pose estimation by synchronizing local unposed nerfs.
\newblock \emph{arXiv preprint arXiv:2306.05410}, 2023.

\bibitem[{Fridovich-Keil and Yu} et~al.(2022){Fridovich-Keil and Yu}, Tancik, Chen, Recht, and Kanazawa]{yu_and_fridovichkeil2021plenoxels}
{Fridovich-Keil and Yu}, Matthew Tancik, Qinhong Chen, Benjamin Recht, and Angjoo Kanazawa.
\newblock Plenoxels: Radiance fields without neural networks.
\newblock In \emph{CVPR}, 2022.

\bibitem[Griwodz et~al.(2021)Griwodz, Gasparini, Calvet, Gurdjos, Castan, Maujean, De~Lillo, and Lanthony]{griwodz2021alicevision}
Carsten Griwodz, Simone Gasparini, Lilian Calvet, Pierre Gurdjos, Fabien Castan, Benoit Maujean, Gregoire De~Lillo, and Yann Lanthony.
\newblock Alicevision meshroom: An open-source 3d reconstruction pipeline.
\newblock In \emph{Proceedings of the 12th ACM Multimedia Systems Conference}, pages 241--247, 2021.

\bibitem[Gu{\'e}don and Lepetit(2023)]{guedon2023sugar}
Antoine Gu{\'e}don and Vincent Lepetit.
\newblock Sugar: Surface-aligned gaussian splatting for efficient 3d mesh reconstruction and high-quality mesh rendering.
\newblock \emph{arXiv preprint arXiv:2311.12775}, 2023.

\bibitem[Jiang et~al.(2020)Jiang, Jiang, and Jiang]{jiang2020efficient}
San Jiang, Cheng Jiang, and Wanshou Jiang.
\newblock Efficient structure from motion for large-scale uav images: A review and a comparison of sfm tools.
\newblock \emph{ISPRS Journal of Photogrammetry and Remote Sensing}, 167:\penalty0 230--251, 2020.

\bibitem[Kerbl et~al.(2023{\natexlab{a}})Kerbl, Kopanas, Leimk{\"u}hler, and Drettakis]{kerbl20233d}
Bernhard Kerbl, Georgios Kopanas, Thomas Leimk{\"u}hler, and George Drettakis.
\newblock 3d gaussian splatting for real-time radiance field rendering.
\newblock \emph{ACM Transactions on Graphics}, 42\penalty0 (4), 2023{\natexlab{a}}.

\bibitem[Kerbl et~al.(2023{\natexlab{b}})Kerbl, Kopanas, Leimk{\"u}hler, and Drettakis]{kerbl3Dgaussians}
Bernhard Kerbl, Georgios Kopanas, Thomas Leimk{\"u}hler, and George Drettakis.
\newblock 3d gaussian splatting for real-time radiance field rendering.
\newblock \emph{ACM Transactions on Graphics}, 42\penalty0 (4), July 2023{\natexlab{b}}.
\newblock URL \url{https://repo-sam.inria.fr/fungraph/3d-gaussian-splatting/}.

\bibitem[Knapitsch et~al.(2017)Knapitsch, Park, Zhou, and Koltun]{Knapitsch2017}
Arno Knapitsch, Jaesik Park, Qian-Yi Zhou, and Vladlen Koltun.
\newblock Tanks and temples: Benchmarking large-scale scene reconstruction.
\newblock \emph{ACM Transactions on Graphics}, 36\penalty0 (4), 2017.

\bibitem[Li et~al.(2022)Li, Tancik, and Kanazawa]{li2022nerfacc}
Ruilong Li, Matthew Tancik, and Angjoo Kanazawa.
\newblock Nerfacc: A general nerf accleration toolbox.
\newblock \emph{arXiv preprint arXiv:2210.04847}, 2022.

\bibitem[Li et~al.(2024)Li, Fidler, Kanazawa, and Williams]{li2024nerfxl}
Ruilong Li, Sanja Fidler, Angjoo Kanazawa, and Francis Williams.
\newblock Nerf-xl: Scaling nerfs with multiple gpus, 2024.

\bibitem[Li et~al.(2023)Li, M{\"u}ller, Evans, Taylor, Unberath, Liu, and Lin]{li2023neuralangelo}
Zhaoshuo Li, Thomas M{\"u}ller, Alex Evans, Russell~H Taylor, Mathias Unberath, Ming-Yu Liu, and Chen-Hsuan Lin.
\newblock Neuralangelo: High-fidelity neural surface reconstruction.
\newblock In \emph{Proceedings of the IEEE/CVF Conference on Computer Vision and Pattern Recognition}, pages 8456--8465, 2023.

\bibitem[Lic{\u{a}}ret et~al.(2022)Lic{\u{a}}ret, Robu, Marcu, Costea, Slu{\c{s}}anschi, Sukthankar, and Leordeanu]{licuaret2022ufo}
Vlad Lic{\u{a}}ret, Victor Robu, Alina Marcu, Drago{\c{s}} Costea, Emil Slu{\c{s}}anschi, Rahul Sukthankar, and Marius Leordeanu.
\newblock Ufo depth: Unsupervised learning with flow-based odometry optimization for metric depth estimation.
\newblock In \emph{2022 International Conference on Robotics and Automation (ICRA)}, pages 6526--6532. IEEE, 2022.

\bibitem[Lin et~al.(2023)Lin, Zhang, Ramanan, and Tulsiani]{lin2023relposepp}
Amy Lin, Jason~Y Zhang, Deva Ramanan, and Shubham Tulsiani.
\newblock Relpose++: Recovering 6d poses from sparse-view observations.
\newblock \emph{arXiv preprint arXiv:2305.04926}, 2023.

\bibitem[Lin et~al.(2021)Lin, Ma, Torralba, and Lucey]{lin2021barf}
Chen-Hsuan Lin, Wei-Chiu Ma, Antonio Torralba, and Simon Lucey.
\newblock Barf: Bundle-adjusting neural radiance fields.
\newblock In \emph{Proceedings of the IEEE/CVF International Conference on Computer Vision}, pages 5741--5751, 2021.

\bibitem[Meuleman et~al.(2023)Meuleman, Liu, Gao, Huang, Kim, Kim, and Kopf]{meuleman2023localrf}
Andreas Meuleman, Yu-Lun Liu, Chen Gao, Jia-Bin Huang, Changil Kim, Min~H. Kim, and Johannes Kopf.
\newblock Progressively optimized local radiance fields for robust view synthesis.
\newblock In \emph{CVPR}, 2023.

\bibitem[Mildenhall et~al.(2021)Mildenhall, Srinivasan, Tancik, Barron, Ramamoorthi, and Ng]{mildenhall2021nerf}
Ben Mildenhall, Pratul~P Srinivasan, Matthew Tancik, Jonathan~T Barron, Ravi Ramamoorthi, and Ren Ng.
\newblock Nerf: Representing scenes as neural radiance fields for view synthesis.
\newblock \emph{Communications of the ACM}, 65\penalty0 (1):\penalty0 99--106, 2021.

\bibitem[M\"uller(2021)]{tinycudann}
Thomas M\"uller.
\newblock {tiny-cuda-nn}, 4 2021.
\newblock URL \url{https://github.com/NVlabs/tiny-cuda-nn}.

\bibitem[M{\"u}ller et~al.(2022)M{\"u}ller, Evans, Schied, and Keller]{muller2022instant}
Thomas M{\"u}ller, Alex Evans, Christoph Schied, and Alexander Keller.
\newblock Instant neural graphics primitives with a multiresolution hash encoding.
\newblock \emph{arXiv preprint arXiv:2201.05989}, 2022.

\bibitem[Ronneberger et~al.(2015)Ronneberger, Fischer, and Brox]{ronneberger2015u}
Olaf Ronneberger, Philipp Fischer, and Thomas Brox.
\newblock U-net: Convolutional networks for biomedical image segmentation.
\newblock In \emph{Medical Image Computing and Computer-Assisted Intervention--MICCAI 2015: 18th International Conference, Munich, Germany, October 5-9, 2015, Proceedings, Part III 18}, pages 234--241. Springer, 2015.

\bibitem[Sch\"{o}nberger and Frahm(2016)]{schoenberger2016sfm}
Johannes~Lutz Sch\"{o}nberger and Jan-Michael Frahm.
\newblock Structure-from-motion revisited.
\newblock In \emph{Conference on Computer Vision and Pattern Recognition (CVPR)}, 2016.

\bibitem[Sch\"{o}nberger et~al.(2016)Sch\"{o}nberger, Zheng, Pollefeys, and Frahm]{schoenberger2016mvs}
Johannes~Lutz Sch\"{o}nberger, Enliang Zheng, Marc Pollefeys, and Jan-Michael Frahm.
\newblock Pixelwise view selection for unstructured multi-view stereo.
\newblock In \emph{European Conference on Computer Vision (ECCV)}, 2016.

\bibitem[Sun et~al.(2022)Sun, Sun, and Chen]{sun2022direct}
Cheng Sun, Min Sun, and Hwann-Tzong Chen.
\newblock Direct voxel grid optimization: Super-fast convergence for radiance fields reconstruction.
\newblock In \emph{Proceedings of the IEEE/CVF Conference on Computer Vision and Pattern Recognition}, pages 5459--5469, 2022.

\bibitem[Suzuki(2024)]{suzuki2024fed3dgs}
Teppei Suzuki.
\newblock {Fed3DGS: Scalable 3D Gaussian Splatting with Federated Learning}.
\newblock \emph{arXiv preprint arXiv:2403.11460}, 2024.

\bibitem[Tancik et~al.(2022)Tancik, Casser, Yan, Pradhan, Mildenhall, Srinivasan, Barron, and Kretzschmar]{tancik2022block}
Matthew Tancik, Vincent Casser, Xinchen Yan, Sabeek Pradhan, Ben Mildenhall, Pratul~P Srinivasan, Jonathan~T Barron, and Henrik Kretzschmar.
\newblock Block-nerf: Scalable large scene neural view synthesis.
\newblock In \emph{Proceedings of the IEEE/CVF Conference on Computer Vision and Pattern Recognition}, pages 8248--8258, 2022.

\bibitem[Tang et~al.(2022)Tang, Zhou, Chen, Hu, Ding, Wang, and Zeng]{tang2022nerf2mesh}
Jiaxiang Tang, Hang Zhou, Xiaokang Chen, Tianshu Hu, Errui Ding, Jingdong Wang, and Gang Zeng.
\newblock Delicate textured mesh recovery from nerf via adaptive surface refinement.
\newblock \emph{arXiv preprint arXiv:2303.02091}, 2022.

\bibitem[Triggs et~al.(2000)Triggs, McLauchlan, Hartley, and Fitzgibbon]{triggs2000bundle}
Bill Triggs, Philip~F McLauchlan, Richard~I Hartley, and Andrew~W Fitzgibbon.
\newblock Bundle adjustment—a modern synthesis.
\newblock In \emph{Vision Algorithms: Theory and Practice: International Workshop on Vision Algorithms}, pages 298--372. Springer, 2000.

\bibitem[Turki et~al.(2022{\natexlab{a}})Turki, Ramanan, and Satyanarayanan]{Turki_2022_CVPR}
Haithem Turki, Deva Ramanan, and Mahadev Satyanarayanan.
\newblock Mega-nerf: Scalable construction of large-scale nerfs for virtual fly-throughs.
\newblock In \emph{Proceedings of the IEEE/CVF Conference on Computer Vision and Pattern Recognition (CVPR)}, pages 12922--12931, June 2022{\natexlab{a}}.

\bibitem[Turki et~al.(2022{\natexlab{b}})Turki, Ramanan, and Satyanarayanan]{turki2022mega}
Haithem Turki, Deva Ramanan, and Mahadev Satyanarayanan.
\newblock Mega-nerf: Scalable construction of large-scale nerfs for virtual fly-throughs.
\newblock In \emph{Proceedings of the IEEE/CVF Conference on Computer Vision and Pattern Recognition}, pages 12922--12931, 2022{\natexlab{b}}.

\bibitem[Turki et~al.(2023)Turki, Zhang, Ferroni, and Ramanan]{turki2023suds}
Haithem Turki, Jason~Y. Zhang, Francesco Ferroni, and Deva Ramanan.
\newblock Suds: Scalable urban dynamic scenes, 2023.

\bibitem[Wang et~al.(2023{\natexlab{a}})Wang, Rupprecht, and Novotny]{wang2023posediffusion}
Jianyuan Wang, Christian Rupprecht, and David Novotny.
\newblock Posediffusion: Solving pose estimation via diffusion-aided bundle adjustment.
\newblock In \emph{Proceedings of the IEEE/CVF International Conference on Computer Vision}, pages 9773--9783, 2023{\natexlab{a}}.

\bibitem[Wang et~al.(2021)Wang, Liu, Liu, Theobalt, Komura, and Wang]{wang2021neus}
Peng Wang, Lingjie Liu, Yuan Liu, Christian Theobalt, Taku Komura, and Wenping Wang.
\newblock Neus: Learning neural implicit surfaces by volume rendering for multi-view reconstruction.
\newblock In \emph{Proc. Advances in Neural Information Processing Systems (NeurIPS)}, volume~34, pages 27171--27183, 2021.

\bibitem[Wang et~al.(2023{\natexlab{b}})Wang, Han, Habermann, Daniilidis, Theobalt, and Liu]{neus2}
Yiming Wang, Qin Han, Marc Habermann, Kostas Daniilidis, Christian Theobalt, and Lingjie Liu.
\newblock Neus2: Fast learning of neural implicit surfaces for multi-view reconstruction.
\newblock In \emph{Proceedings of the IEEE/CVF International Conference on Computer Vision (ICCV)}, 2023{\natexlab{b}}.

\bibitem[Xu et~al.(2023)Xu, Xiangli, Peng, Pan, Zhao, Theobalt, Dai, and Lin]{xu2023gridguided}
Linning Xu, Yuanbo Xiangli, Sida Peng, Xingang Pan, Nanxuan Zhao, Christian Theobalt, Bo~Dai, and Dahua Lin.
\newblock Grid-guided neural radiance fields for large urban scenes, 2023.

\bibitem[Yao et~al.(2020)Yao, Luo, Li, Zhang, Ren, Zhou, Fang, and Quan]{yao2020blendedmvs}
Yao Yao, Zixin Luo, Shiwei Li, Jingyang Zhang, Yufan Ren, Lei Zhou, Tian Fang, and Long Quan.
\newblock Blendedmvs: A large-scale dataset for generalized multi-view stereo networks, 2020.

\bibitem[Zhang et~al.(2020)Zhang, Riegler, Snavely, and Koltun]{zhang2020nerf++}
Kai Zhang, Gernot Riegler, Noah Snavely, and Vladlen Koltun.
\newblock Nerf++: Analyzing and improving neural radiance fields.
\newblock \emph{arXiv preprint arXiv:2010.07492}, 2020.

\bibitem[Zhang et~al.(2023)Zhang, Kundu, Funkhouser, Guibas, Su, and Genova]{zhang2023nerflets}
Xiaoshuai Zhang, Abhijit Kundu, Thomas Funkhouser, Leonidas Guibas, Hao Su, and Kyle Genova.
\newblock Nerflets: Local radiance fields for efficient structure-aware 3d scene representation from 2d supervisio.
\newblock \emph{arXiv preprint arXiv:2303.03361}, 2023.

\bibitem[Zhao et~al.(2023)Zhao, Gou, Li, Peng, Lv, and Peng]{zhao2023comprehensive}
Haiyu Zhao, Yuanbiao Gou, Boyun Li, Dezhong Peng, Jiancheng Lv, and Xi~Peng.
\newblock Comprehensive and delicate: An efficient transformer for image restoration.
\newblock In \emph{Proceedings of the IEEE/CVF Conference on Computer Vision and Pattern Recognition}, pages 14122--14132, 2023.

\end{thebibliography}

\end{document}